\documentclass{article}
\usepackage{tcolorbox}
\usepackage{listings}
\usepackage[table]{xcolor}
\usepackage{amsmath}
\usepackage{booktabs}
\usepackage{spconf,amsmath,graphicx,hyperref}

\title{TennisTV: Do Multimodal Large Language Models Understand Tennis Rallies?}
\twoauthors
  {Zhongyuan Bao\textsuperscript{*}\thanks{*Equal Contribution.}}
	{Independent Researcher\\
	Shanghai, China}
  {Lejun Zhang\textsuperscript{*}}
	{New York University\\
	Tandon School of Engineering\\
	New York, USA}
    
\begin{document}
\ninept
\maketitle
\begin{abstract}

Multimodal large language models (MLLMs) excel at general video understanding but struggle with fast, high-frequency sports like tennis, where rally clips are short yet information-dense. To systematically evaluate MLLMs in this challenging domain, we present TennisTV, the first and most comprehensive benchmark for tennis video understanding. TennisTV models each rally as a temporal-ordered sequence of consecutive stroke events, using automated pipelines for filtering and question generation. It covers 8 tasks from the stroke-level to the rally-level and includes 2527 human-verified questions. Evaluating 17 representative MLLMs, we provide the first systematic assessment of tennis video understanding. Results yield two key insights: (i) frame-sampling density should be tailored and balanced across tasks, and (ii) improving temporal grounding is essential for stronger reasoning. Our dataset can be found in this \href{https://modelscope.cn/datasets/FDUBay/TennisTV}{url}.

\end{abstract}
\section{Introduction}
\label{sec:intro}

\begin{figure}[ht]
    \centering
    \includegraphics[width=1\linewidth]{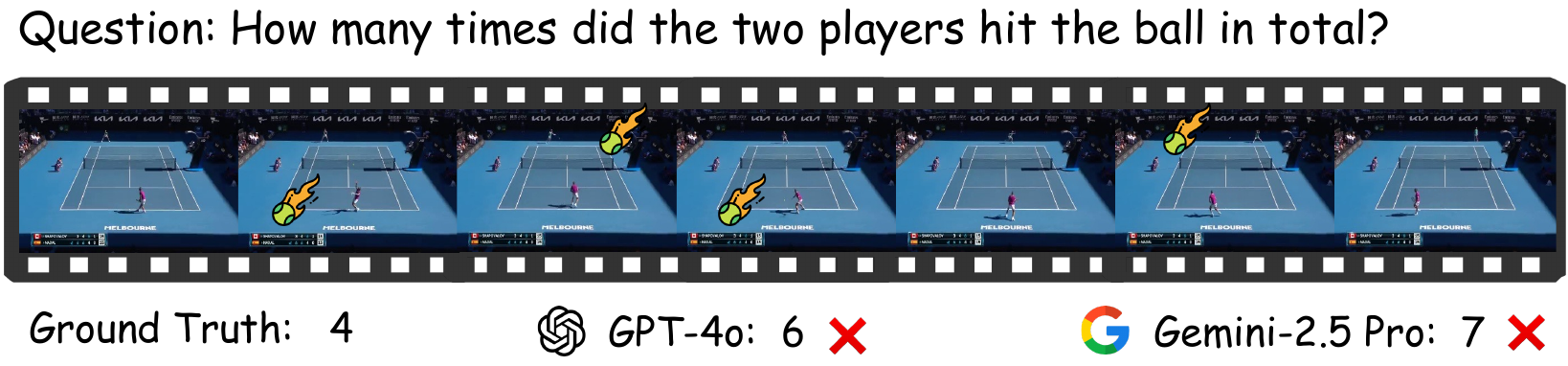}
    \caption{Illustrations of a simple counting task easily handled by non-experts, but still challenging for the advanced commercial models. }
    \label{fig:tennis_teaser}
\end{figure}

Tennis, renowned worldwide for its commercial impact and elite tournaments, has recently attracted significant research interest in applying artificial intelligence to tennis video understanding. However, despite the significant progress made in general video understanding\cite{maaz2023video,wang2025internvideo2} and reasoning\cite{li2025videochat,wang2025videorft}, identifying fast-paced, high-frequency sports videos is still challenging\cite{liu2025f}. As shown in the Fig~\ref{fig:tennis_teaser}, for human observers, even in dynamic scenes, they can directly depict the start and end of each stroke, infer the type of stroke and the trajectory of the ball. In contrast, even the most advanced MLLMs are struggling to deal with seemingly basic tasks, such as calculating the total number of hits in a rally.

In traditional video understanding, short videos typically require fewer frames due to their limited duration, whereas long videos generally rely on denser frame sampling. In contrast, tennis rally videos, though brief in duration, are highly information-dense. A single rally lasting fewer than 10 seconds may contain high-density, rapidly changing motion signals, such as a player’s stroke type choice, movement trajectory, the ball's rebound position etc. The central challenge is that, despite their brevity, the frequent and high-speed dynamics of rallies demand dense frame sampling, often approaching the level required for long video analysis; otherwise, crucial hitting details are likely to be lost.

% 为了系统性的评估当前MLLMs对于这类short, information-dense体育视频的理解能，as well as 推进社区对于MLLMs 在网球视频理解能力上的提升，我们推出了 TennisTV，这是一个专为理解网球回合视频而设计的高质量基准测试。TennisTV在概念上将每个网球回合视频解构为了一条由若干个时序上连续的击球事件组成的rally sequence，共设计了8个契合网球的运动特点的评估任务，包括多项选择题与判断题两种题型。在benchmark构造上，我们设计了一套自动化的annotation pipeline，通过三个阶段的标注流程，实现了在原始数据源提供的粗糙标注的基础上对于整个回合进行结构化的注释整合，额外选手运动信息的抽取，以及问题的生成。最终得到 2500条经过人工质量审核的TennisTV benchmark。

In order to systematically evaluate the current MLLMs' understanding of short, information-dense sports videos, as well as promote the community's improvement of MLLMs' understanding of tennis videos, we introduce TennisTV, the first high-quality benchmark specifically designed for understanding tennis rally videos. TennisTV defines a stroke event as the atomic unit, spanning movement initiation, stroke action choice, directional execution, and the ball’s landing. Each rally video is modeled as a time-ordered rally sequence composed of these consecutive stroke events. A total of 8 evaluation tasks that match the sports characteristics of tennis, including two types of multiple-choice questions and judgment statements. We design an automated annotation pipeline. Through three-stage annotation process, we realize structured annotation integration for the entire rally based on the rough annotation provided by the original data source\cite{liu2025f}, extract additional athletes' movement information, and generate questions. Finally, through manual quality control, we construct the TennisTV benchmark, which contains 2527 curated questions.

%【替换下面这段】我们采用了17个先进的前沿的MLLMs在TennisTV上进行了全面的评估与分析，包括3个闭源商用模型，8个开源non-thinking VideoLLM以及5个开源Thinking VideoLLM。在实验中，我们发现，三个findings (i)部分tasks存在high degree of correlation.(ii)Optimal frame sampling density differs by task(iii)hinking brings gains on knowledge-light tasks。基于这些insights, 提出两个takeaways:（i）平衡好frame sampling density和任务粒度（ii）继续探索thinking model带来的提升。We hope TennisTV and the above insights will guide future advances in tennis video understanding.

We conduct a comprehensive evaluation of 17 state of the art MLLMs on TennisTV, including three closed-source commercial models, eight open-source non-thinking VideoLLMs, and five open-source thinking VideoLLMs. The study yields three findings: first, certain tasks exhibit a high degree of correlation; second, the optimal frame sampling density differs across tasks; third, explicit reasoning delivers gains on knowledge light tasks. From these insights above, we draw two practical takeaways: balance frame sampling density with task granularity, and to further enhance the temporal grounding capabilities of reasoning models. We hope that TennisTV and these insights will guide future advances in tennis video understanding.

\begin{figure*}[ht]
    \centering
    \includegraphics[width=0.9\linewidth]{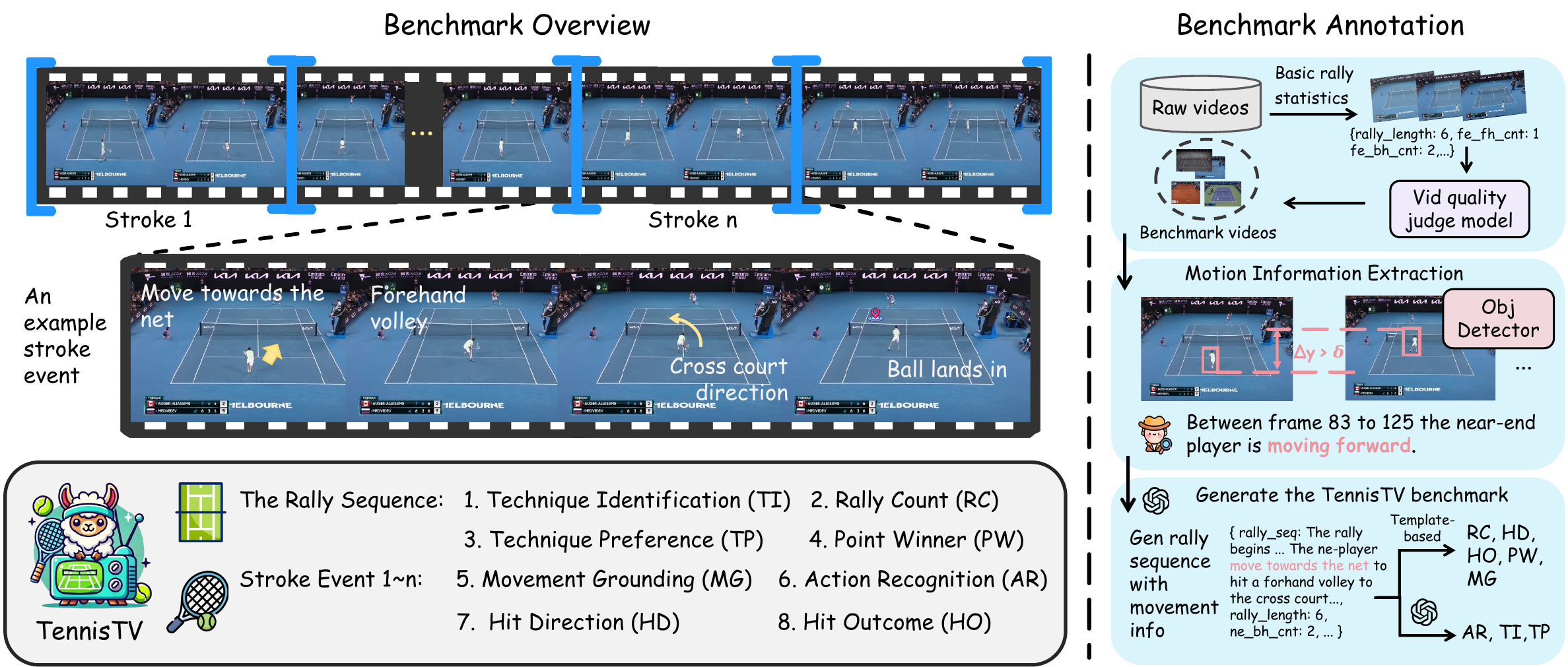}
    \caption{Overview and annotation pipeline of TennisTV Benchmark.}
    \label{fig:tennis_bench}
\end{figure*}

\section{Related Work}
\label{sec:format}

\subsection{Sports Understanding}
Sports understanding \cite{thomas2017computer} is a rapidly evolving field that encompasses multiple research topics and integrates diverse modalities, covering tasks such as action recognition \cite{deliege2021soccernet}, athlete analysis \cite{shao2020finegym}, tactical planning \cite{liu2023insight}, foul recognition \cite{felsen2017will}. Recent work on MLLMs \cite{li2024sports,xia2024sportu,xia2024sportqa} has shifted toward general frameworks spanning diverse sports tasks, while existing work\cite{liu2025f} in tennis primarily uses conventional neural networks for stroke label prediction, with limited modeling of rally-level context and causal relations. Consequently, MLLM capabilities in tennis video remain largely unexplored. This paper uses tennis as the empirical domain and proposes a novel benchmark for fast-paced, high-frequency sports video understanding, with the goal of advancing future research in this area.

\subsection{MLLMs for Video}
Inspired by the success of LLMs in natural language processing, researchers have begun to explore their extension to multimodal inputs, leading to the development of MLLMs. In image domain, \cite{li2023blip,liu2023visual} have demonstrated that integrating visual encoders with LLMs can effectively process image-text pairs. However, video understanding presents greater challenges due to the temporal information\cite{zou2024seconds}. Some early approaches extended LLMs to video domain by incorporating temporal modeling mechanisms\cite{lin2023video}. Recent research has increasingly focused on improving the temporal and regional comprehension capabilities of Video MLLMs. With the emergence of Deepseek-R1\cite{guo2025deepseek}, a growing body of work has begun to apply reinforcement learning (RL) to substantially enhance the reasoning abilities of MLLMs\cite{li2025videochat,chen2025versavid,wang2025videorft}, thereby enabling stronger video understanding. However, little attention has been given to investigating how Video MLLMs interpret short, information-dense sports events. To address this gap, we conduct experiments using the carefully designed TennisTV to evaluate the benefits of enhancing base models for analyzing fast-pace, high-frequency sports videos.

\begin{table*}[ht]
\small
\centering
\setlength{\tabcolsep}{4pt}
\caption{Main results on TennisTV. AvgSE and AvgRS denote the average metrics of tasks related to stroke event and rally sequence.}
\label{tab:benchmarking}
\begin{tabular}{lccccc|c|cccc|c|c}
\toprule
\textbf{Method} & \textbf{Frame Sampling} & \textbf{MG} & \textbf{AR} & \textbf{HD} & \textbf{HO} & \textbf{AvgSE} & \textbf{TP} & \textbf{RC} & \textbf{TI} & \textbf{PW} &  \textbf{AvgRS} & \textbf{Avg} \\
\midrule
Random & - & 20.0 & 25.0 & 20.0 & 33.3 & 24.6 & 25.0 & 25.0 & 25.0 & 50.0 & 31.3 & 	27.9  \\
Human-level & - & 47.8 & 100.0 & 96.3 & 93.3 & 84.4 & 91.6 & 92.5 & 98.0 & 94.3 & 94.1   & 	89.2 \\
\midrule
\rowcolor{gray!20}
\multicolumn{13}{l}{\textit{Closed-Source Models}} \\
Gemini 2.5 Pro & 1 FPS & \textbf{41.7} & 47.4 & \underline{34.0} & 45.0 & \underline{42.0} & \underline{49.1} & 31.5 & \textbf{51.0} & 51.7 & 45.8   & 43.9  \\
GPT-4.1 & 1 FPS & \underline{23.3} & \textbf{59.7} & 32.5 & \textbf{52.7} & \textbf{42.1} & \textbf{51.6} & \underline{38.8} & 49.3 & \underline{53.0} &  \underline{48.2} & 	\textbf{45.1}  \\
GPT-4o & 1 FPS & 18.3 & \underline{56.6} & \textbf{34.8} & \underline{52.3} & 40.5 & 47.5 & \textbf{39.8} & \underline{50.3} & \textbf{56.3} &   \textbf{48.5} 	& \underline{44.5} \\
\midrule
\rowcolor{gray!20}
\multicolumn{13}{l}{\textit{Open-Source Models}} \\
Video-LLaVA-7B & 8 Frames & 19.4 & 25.2 & 16.3 & 32.3 & 23.3 & 24.8 & 23.3 & 25.0 & 50.0  &   30.8 & 	27.0   \\
LLaVA-OV-7B & 64 Frames & 14.4 & 36.3 & 30.0 & 41.0 & 30.4 & 35.7 & 25.3 & 41.0 & 49.0  &  37.8 & 	34.1   \\
mPLUG-Owl3 & 128 Frames & 16.1 & 42.8 & 33.8 & \textbf{46.0} & 34.7 & 35.7 & 18.3 & 40.7 & 50.0  &  36.2 & 	35.4   \\
Qwen2VL-7B & 32 Frames & 19.4 & 29.5 & 34.5 & 40.3 & 30.9 & \underline{43.5} & 24.8 & 44.7 & 50.0 &  40.8 & 	35.8  \\
InternVideo2.5-8B & 128 Frames & 20.0 & \underline{49.5} & 17.8 & 43.0 & 32.6 & 42.5 & 19.0 & 43.3 & 50.3 &  38.8 & 	35.7 \\
InternVL3-8B & 64 Frames & \textbf{31.9} & 34.8 & 31.3 & 45.3 & 35.8 & 40.7 & 30.3 & 44.3 & 50.0 & \underline{41.3} & 	38.5   \\
Qwen2.5VL-3B & 32 Frames & 17.8 & 39.7 & 32.0 & 40.3 & 32.5 & 39.8 & 20.5 & 46.3 & 50.0 &   39.2 & 	35.8   \\
Qwen2.5VL-7B & 32 Frames & 26.7 & 48.6 & 34.5 & \underline{45.7} & \underline{38.9} & 42.2 & 22.3 & 42.7 & 48.0 & 38.8   & 38.8  \\
MiMoVL-7B SFT & 32 Frames & 22.8 & 45.5 & 31.8 & 38.3 & 34.6 & \textbf{44.7} & \underline{31.8} & 41.0 & 46.0 &  40.9   & 	37.4 \\
\midrule
\rowcolor{gray!20}
\multicolumn{13}{l}{\textit{Open-Source Thinking Models}} \\
Video-R1 & 32 Frames & 21.7 & 45.9 & 28.8 & 42.0 & 34.6 & 39.4 & 28.3 & 40.7 & \underline{52.0}   & 40.1 & 	36.8  \\
VCR-Thinking-7B & 32 Frames & \underline{28.3} & 45.5 & \underline{35.5} & 23.7 & 33.3 & 38.5 & \textbf{33.3} & 44.0 & 47.7 &  40.9 &	38.9  \\
VideoRFT & 32 Frames & \underline{28.3} & 46.8 & 35.3 & 45.0 & \underline{38.9} & 37.3 & 25.5 & 45.3 & 45.7   &  38.5 &	38.1  \\
VersaVid-R1 & 32 Frames & 24.4 & \textbf{52.3} & \textbf{36.8} & 44.7 & \textbf{39.6} & 37.9 & 27.0 & \underline{46.7} & 50.0   &  40.4 &	\textbf{40.6}   \\
MiMoVL-7B RL & 32 Frames & 22.8 & 48.0 & 33.3 & 42.7 & 36.7 & 36.7 & 28.8 & \textbf{49.3} & \textbf{57.3}   & \textbf{44.3}   & 	\underline{40.5} \\
\bottomrule
\end{tabular}
\end{table*}

\section{TennisTV Benchmark}
To rigorously evaluate the performance of MLLMs in tennis video understanding, we introduce \textbf{Tennis} \textbf{T}our \textbf{V}ideo (TennisTV). The benchmark comprises 8 subtasks with 2527 questions, supporting multiple levels of video understanding. This section outlines benchmark introduction and automatic annotation pipeline.

\begin{figure}[ht]
    \centering
    \includegraphics[width=0.9\linewidth]{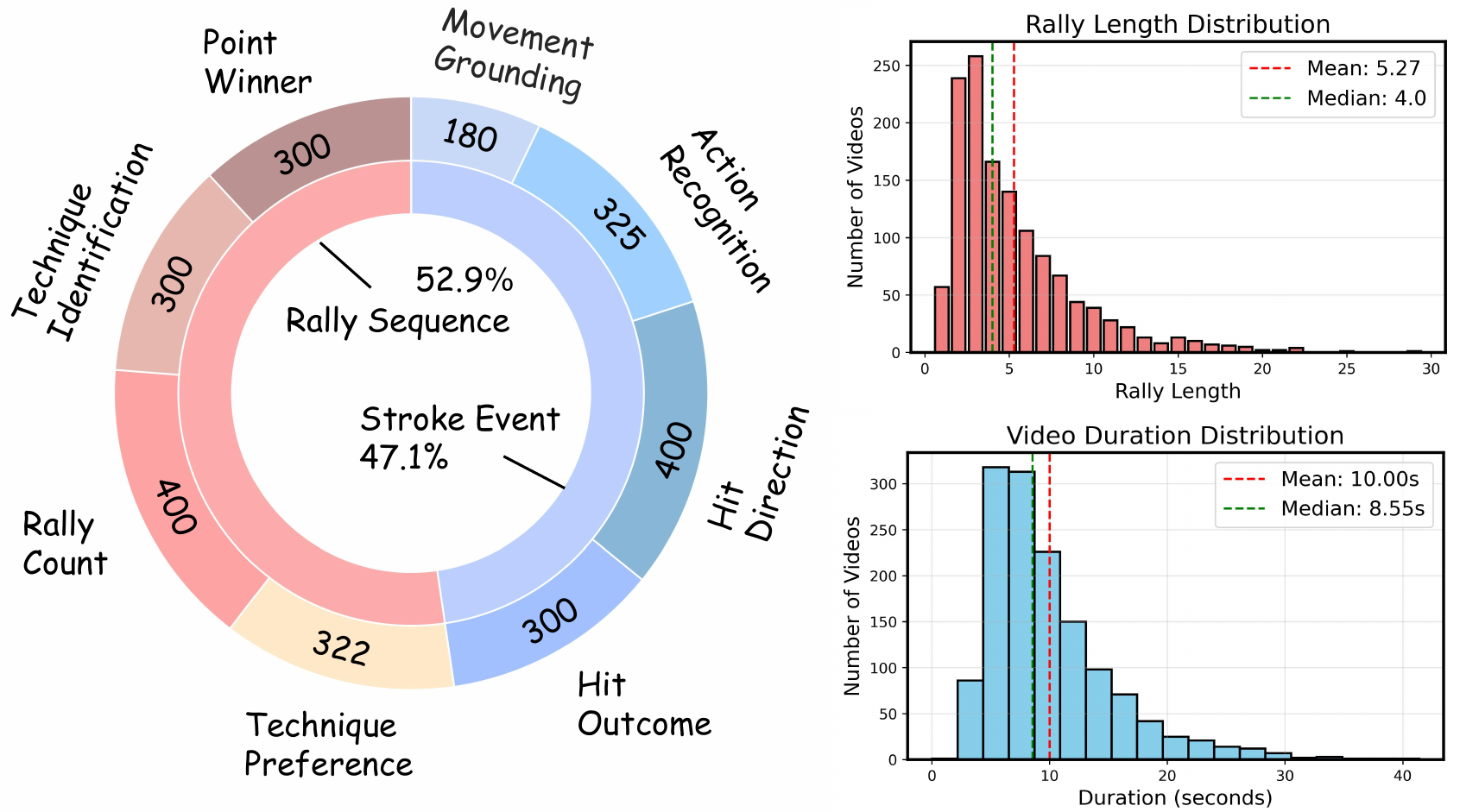}
    \caption{Benchmark statistics of TennisTV.}
    \label{fig:bench_stat}
\end{figure}

\subsection{Benchmark Task Analysis}
%当我们观看一场网球比赛的视频片段时，看到的是一段高速、高频的技战术交锋。每一回合由一方运动员发起，完成启动移动、选择击球方式、选择击球线路，并最终产生具体的击球落点或结果，球过网或反弹后，对方选手接续执行相同的流程。这样的事件序列在双方之间循环交替，直至某一次击球的结果导致回合终结，构成一个完整且连贯的回合过程。基于上述对于这项运动本质的分析，我们将每一回合的胜负层进行了层次化的拆分。一次得分是由rally sequence的结果决定的，同时，每一个rally sequence是由n个stroke event 组成的。Fine- grained的stroke event推动着rally sequence的走势变化，最终rally sequence的结果决定了哪位运动员获得了这一分。
When viewing a tennis rally, we observe a rapid, high-frequency exchange of tactics and technique. Each rally is initiated by one player, who accelerates into position, selects a stroke type, chooses a shot direction, and ultimately produces a specific landing point or outcome. After the ball crosses the net or rebounds, the opponent proceeds through the same process. The event sequence alternates between players until a stroke ends the point. Based on this understanding of the sport, TennisTV defines a stroke event as the atomic unit, spanning movement initiation, shot selection, directional execution, and ball landing. Each rally video is modeled as a time-ordered rally sequence composed of these consecutive stroke events.

\noindent\textbf{Stroke Event.} The stroke-level perspective addresses fine-grained details. Action Recognition (AR) requires the model to classify basic stroke types, while Movement Grounding (MG) targets the precise temporal localization of player movements. Moreover, Hit Direction (HD)  emphasizes on the ball's trajectory, and Hit Outcome (HO) asks the model to determine the immediate result of one stroke. These tasks jointly depend on the model's ability to capture subtle spatiotemporal cues at the frame level.

\noindent\textbf{Rally Sequence.} The rally-level perspective captures the macro dynamics of a rally. Technique Identification (TI) requires the model to detect specific tactics, while Rally Count (RC) tasks it with recording the total number of strokes. Technique Preference (TP) demands that the model analyze in-rally tactical choices to reveal player tendencies, and Point Winner (PW) requires the model to identify the winner of the point. These tasks collectively necessitate a holistic analysis to capture overarching technical patterns.

Overall, TennisTV offers a comprehensive benchmark that probes both basic perception (e.g., AR, HD) and tactical reasoning (e.g., PW, TP), providing a robust, diversified testbed for video understanding in fast-paced, high-frequency sports.

\subsection{Dataset Annotation Pipeline}
TennisTV is derived from F3Set\cite{liu2025f}, which provides manual annotations of individual shots from tennis tournament videos. Based on these annotations, we perform a three-stage annotation pipeline.

\noindent\textbf{Basic Data Preparation} 
%我们首先对于属于同一个rally的击球注释进行时序上的汇总与基础的回合技战术统计（包含回合长度，双方选手使用的不同动作统计等），适当去除原始数据中回合数量过少的视频与标注，得到结构化的回合信息。此外，由于网球比赛中的场地类型五花八门，某些场地背景颜色会导致观感下降，因此，我们引入qwen2.5VL-7b作为视频质量审核模型，排除那些观感体验不清晰的视频。
We first aggregated the shot annotations corresponding to the same rally in temporal order and generated basic rally-level technical statistics, such as rally length and the distribution of different strokes used by both players, to construct structured rally information. Next, we removed videos with an insufficient number of rallies from the original dataset. Moreover, because tennis tournaments are played on courts with diverse surfaces, the varying background colors can reduce viewing clarity. To address this issue, we employed Qwen2.5VL-7B\cite{bai2025qwen2} as a video quality judge model to filter out videos with poor visual quality.

\noindent\textbf{Motion Information Extraction} 
%其次，原始的数据标注仅粗糙的关注了单次stroke event中球的飞行轨迹，缺失在整个rally中对于选手移动方向与变化的关注。为此，我们设计了选手运动信息提取方法获取在stroke event之间球员的移动信息。给定一个视频，我们首先从第一阶段汇总过后结构化回合信息，提取原始标注的m个击球帧，相邻两帧进行配对，得到m-1个击球帧对。在一个击球帧对中，我们使用了一个light-weight object detector，对于运动员位置进行grounding，通过对两帧的bounding box坐标 在x，y坐标系上进行delta x，delta y偏移量的计算，当某一方向的偏移量大于threshold时，我们就判断运动员在这两帧之间在这一方向进行了移动。我们获取到的选手运动信息页加入到了结构化回合信息中。
Secondly, the original annotations primarily capture the ball’s trajectory within individual stroke events, while overlooking the player’s movement direction and positional changes throughout the rally. To address this limitation, we developed a player motion extraction method to capture movement information between stroke events. For each video, we extracted $m$ originally annotated stroke frames from the structured rally information obtained in the first stage. We then paired adjacent frames to form $m-1$ stroke frame pairs. For each pair, we applied a lightweight object detector to localize the player’s position. By computing the $\Delta x$ and $\Delta y$ offsets between the bounding box coordinates across the two frames, we determined movement direction: if either offset exceeded a specific threshold \(\delta\), the player is considered to have moved in that direction. Finally, the extracted player motion information was integrated into the structured rally information.

\noindent\textbf{Question Generation} 
%最后，我们使用GPT-4o对于每一个rally的结构化信息进行汇总，得到了一条完整的rally sequence的描述。基于rally sequence以及一些基础的回合统计信息，我们分别基于模版和GPT-4o生成了8个任务的问题。经过人工筛查后，共计产出2500条benchmark的问题。
We employed GPT-4o to aggregate the structured rally information and generated a comprehensive rally sequence description for each video. Using the rally descriptions and basic statistics, we generated 8 task questions using either GPT-4o or a predefined template. All questions were manually verified, resulting in a total of 2527 questions across 1298 videos.

\section{Experiment}
\subsection{Implementation Details}
We evaluate 14 open-source MLLMs that supports multi-frame inputs, including 9 non-thinking models: Video-LLaVA \cite{lin2023video}, LLaVA-OV-7B, Qwen2VL-7B, Qwen2.5VL-3B\&7B \cite{bai2025qwen2}, mPLUG-Owl3 \cite{ye2024mplug}, InternVideo2.5 \cite{wang2025internvideo2}, InternVL3-8B and MiMoVL-SFT-7B-2508, as well as 5 thinking models: Video-R1 \cite{feng2025video} ,VideoChat-R1 \cite{li2025videochat} ,VideoRFT \cite{wang2025videorft} ,VersaVid-R1 \cite{chen2025versavid} and MiMoVL-RL-7B-2508 \cite{xiaomi2025mimo}. We assess GPT-4o, GPT-4.1, Gemini2.5-Pro through their APIs. For open-source models, we followed the input frame settings in VideoMME \cite{fu2025video}. For closed-source models, we adopted 1 FPS video sampling rate for evaluation due to budget consideration.

\subsection{Evaluation Metric}
We assess performance on TennisTV using two formats. In PW task, the model must correctly classify a statement about the rally outcome as True or False. All remaining tasks are cast as multiple-choice questions, evaluated based on the selection of the correct answer.

\subsection{Main Results}

\noindent\textbf{Results}  
Table~\ref{tab:benchmarking} shows results on TennisTV.  Among closed-source models, GPT-4.1 rank first with an overall score of 45.1\%, narrowly surpassing GPT-4o and Gemini 2.5 Pro. Open-source systems also demonstrate strong competitiveness: Qwen2.5-VL-7B achieves a score of 38.8\%, while the best thinking model VersaVid-R1 improves this to 40.6\% (+1.8pp). On average, open-source reasoning models outperform their non-reasoning counterparts but still lag considerably behind leading closed-source commercial systems. Despite these gains, overall accuracy remains well below human performance and even falls beneath the random baseline on some tasks, indicating substantial headroom for progress.
\begin{figure}[htb]
    \centering
    \includegraphics[width=.7\columnwidth]{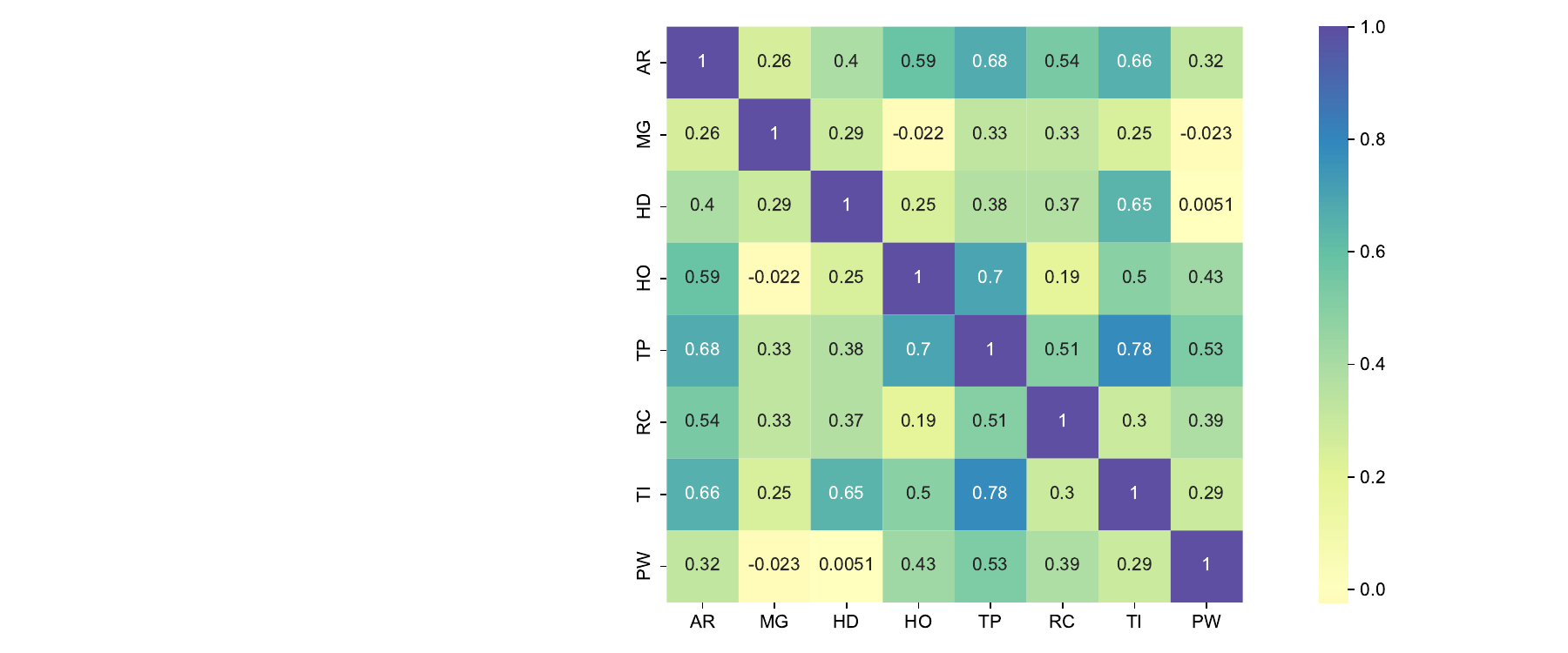}
    \caption{Correlation matrix between 8 tasks in TennisTV.}
    \label{fig:task_correlation}
\end{figure}

\noindent\textbf{Finding I --- High degree of correlation among the AR, TI, and TP tasks.}
Fig~\ref{fig:task_correlation} reveals a robust positive correlation among AR, TI, and TP that persists across diverse model families and sampling protocols. We attribute this to the fact that these tasks form a coherent capability hierarchy: recognizing specific actions yields the primitives for technique identification, which subsequently supports the inference of tactic preferences. Consequently, performance on one task is highly predictive of the others, suggesting that low-level refinements in attribution or temporal stitching can transfer to tactic inference. Taken together, these outcomes underscore the empirical soundness of our benchmark.

\begin{table}[h]
% \small
\centering
\setlength{\tabcolsep}{6.5pt} % 调整列间距
\caption{Ablation study of different frame sampling strategies on Qwen2.5VL-7B. We report the AvgSE, AvgRS and Avg metrics.}
\label{tab:frame_diff}
\begin{tabular}{lccc}
\toprule
 \textbf{Frame Sampling} & \textbf{AvgSE}  & \textbf{AvgRS} & \textbf{Avg} \\
\midrule
 1 FPS &  37.9 & 38.3 & 38.1  \\
 16 Frames & 38.0 & \textbf{39.3} & \underline{38.6} \\
 32 Frames & \underline{38.9}  &  \underline{38.8}  &  \textbf{38.8}  \\
 64 Frames & \textbf{39.1} & 37.7 & 38.4  \\
\toprule
\end{tabular}
\end{table}

\noindent\textbf{Finding II --- The optimal frame sampling density varies across perceptual levels.} In fast-paced, high-frequency sports videos, the number of frames required varies across different levels of understanding. As shown in Table \ref{tab:frame_diff}, performance with Qwen2.5-VL-7B peaks at 32 frames, while adding more frames yields diminishing or even negative returns, particularly for rally-sequence metrics. These results suggest that finer temporal resolution is most beneficial for capturing brief, high-frequency stroke cues, whereas rally-level reasoning saturates earlier and is more sensitive to redundancy.

\begin{figure}[htb]
    \centering
    \includegraphics[width=0.95\linewidth]{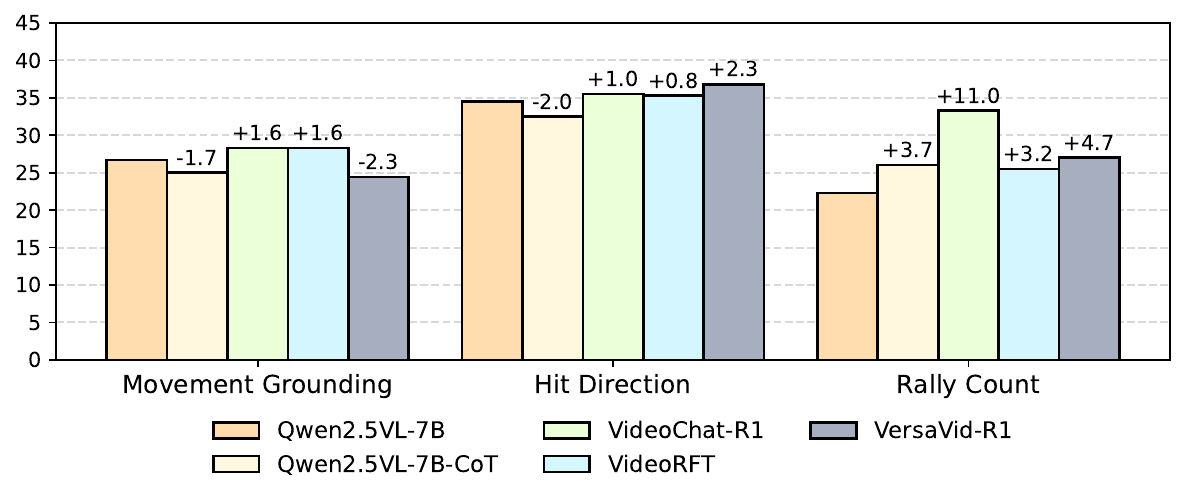}
    \caption{Visualization of performance on three sub-tasks, which require no prior knowledge of tennis, based on the same base model.}
    \label{fig:thinking_ability}
\end{figure}

\noindent\textbf{Finding III --- Reasoning not always helps.}
To disentangle the effect of reasoning from tennis-specific priors, we analyze three knowledge-light tasks (MG, HD, RC) that minimize dependence on specialized rules. We analyze Qwen2.5-VL-7B, the zero-shot chain-of-thought (CoT) prompting method \cite{kojima2022large} and three RL post-trained reasoning variants (VideoChat-R1, VideoRFT, VersaVid-R1). As shown in Fig.~\ref{fig:thinking_ability}, reasoning variants exhibit consistent gains over the non-reasoning baseline on RC and HD, where multi-step aggregation and explicit trajectory reasoning are critical. However, on MG task, which focuses on temporal grounding, reasoning model do not show consistent improvement over the baseline. This suggests that existing RL strategies primarily enhance general content understanding, while temporal perception remains a key shortcoming.

\section{Takeaways}
Based on our findings, we highlight two takeaways
for future improvements in tennis video understanding:

\begin{itemize}
  \item \textbf{Balancing frame sampling density across tasks.}
  For tennis video understanding, the sampling frequency that yields the best performance differs between coarse global tasks and fine-grained tasks. Future work should pursue a principled balance between event granularity and sampling density, so that models capture long range context without losing the detailed temporal cues required for rapid event recognition.

  \item \textbf{Enhancing the temporal understanding ability of video reasoning models.}
 The experimental results have proved that reasoning can improve rally-level understanding, but it has not benefited from fine-grained temporal grounding task. This suggests that future RL post-training should emphasize temporal understanding and alignment mechanisms rather than relying solely on general video reasoning.
\end{itemize}

\section{Conclusion}
In this paper, we present \textbf{TennisTV}, the first benchmark ever for tennis video understanding tailored to evaluate MLLMs. TennisTV models each rally as a time-ordered sequence of consecutive strokes, using automated pipelines for question generation. Evaluations over diverse MLLMs show that reasoning and careful frame sampling help, yet models still struggle with the sport’s fast, information-dense dynamics. We expect TennisTV to spur advances in dynamic sports video understanding.

\clearpage 

%\section{REFERENCES}
% \label{sec:refs}

% List and number all bibliographical references at the end of the
% paper. The references can be numbered in alphabetic order or in
% order of appearance in the document. When referring to them in
% the text, type the corresponding reference number in square
% brackets as shown at the end of this sentence \cite{C2}. An
% additional final page (the fifth page, in most cases) is
% allowed, but must contain only references to the prior
% literature.

% Please follow the IEEE Citation Guidelines, \url{https://ieee-dataport.org/sites/default/files/analysis/27/IEEE\%20Citation\%20Guidelines.pdf} for formatting of references.

% References should be produced using the bibtex program from suitable
% BiBTeX files (here: strings, refs, manuals). The IEEEbib.bst bibliography
% style file from IEEE produces unsorted bibliography list.
% --------------------------------------------
  
{\small
\bibliographystyle{IEEEbib}
\bibliography{strings,refs}

@inproceedings{deliege2021soccernet,
  title={Soccernet-v2: A dataset and benchmarks for holistic understanding of broadcast soccer videos},
  author={Deliege, Adrien and Cioppa, Anthony and Giancola, Silvio and Seikavandi, Meisam J and Dueholm, Jacob V and Nasrollahi, Kamal and Ghanem, Bernard and Moeslund, Thomas B and Van Droogenbroeck, Marc},
  booktitle={Proceedings of the IEEE/CVF conference on computer vision and pattern recognition},
  pages={4508--4519},
  year={2021}
}

@inproceedings{shao2020finegym,
  title={Finegym: A hierarchical video dataset for fine-grained action understanding},
  author={Shao, Dian and Zhao, Yue and Dai, Bo and Lin, Dahua},
  booktitle={Proceedings of the IEEE/CVF conference on computer vision and pattern recognition},
  pages={2616--2625},
  year={2020}
}

@inproceedings{liu2023insight,
  title={Insight analysis for tennis strategy and tactics},
  author={Liu, Zhaoyu and Jiang, Kan and Hou, Zhe and Lin, Yun and Dong, Jin Song},
  booktitle={2023 IEEE International Conference on Data Mining (ICDM)},
  pages={1169--1174},
  year={2023},
  organization={IEEE}
}

@inproceedings{felsen2017will,
  title={What will happen next? forecasting player moves in sports videos},
  author={Felsen, Panna and Agrawal, Pulkit and Malik, Jitendra},
  booktitle={Proceedings of the IEEE international conference on computer vision},
  pages={3342--3351},
  year={2017}
}

@inproceedings{li2023blip,
  title={Blip-2: Bootstrapping language-image pre-training with frozen image encoders and large language models},
  author={Li, Junnan and Li, Dongxu and Savarese, Silvio and Hoi, Steven},
  booktitle={International conference on machine learning},
  pages={19730--19742},
  year={2023},
  organization={PMLR}
}

@inproceedings{fu2025video,
  title={Video-mme: The first-ever comprehensive evaluation benchmark of multi-modal llms in video analysis},
  author={Fu, Chaoyou and Dai, Yuhan and Luo, Yongdong and Li, Lei and Ren, Shuhuai and Zhang, Renrui and Wang, Zihan and Zhou, Chenyu and Shen, Yunhang and Zhang, Mengdan and others},
  booktitle={Proceedings of the Computer Vision and Pattern Recognition Conference},
  pages={24108--24118},
  year={2025}
}

@article{thomas2017computer,
  title={Computer vision for sports: Current applications and research topics},
  author={Thomas, Graham and Gade, Rikke and Moeslund, Thomas B and Carr, Peter and Hilton, Adrian},
  journal={Computer Vision and Image Understanding},
  volume={159},
  pages={3--18},
  year={2017},
  publisher={Elsevier}
}

@article{li2024sports,
  title={Sports-qa: A large-scale video question answering benchmark for complex and professional sports},
  author={Li, Haopeng and Deng, Andong and Ke, Qiuhong and Liu, Jun and Rahmani, Hossein and Guo, Yulan and Schiele, Bernt and Chen, Chen},
  journal={arXiv preprint arXiv:2401.01505},
  year={2024}
}

@article{xia2024sportqa,
  title={Sportqa: A benchmark for sports understanding in large language models},
  author={Xia, Haotian and Yang, Zhengbang and Wang, Yuqing and Tracy, Rhys and Zhao, Yun and Huang, Dongdong and Chen, Zezhi and Zhu, Yan and Wang, Yuan-fang and Shen, Weining},
  journal={arXiv preprint arXiv:2402.15862},
  year={2024}
}

@article{xia2024sportu,
  title={Sportu: A comprehensive sports understanding benchmark for multimodal large language models},
  author={Xia, Haotian and Yang, Zhengbang and Zou, Junbo and Tracy, Rhys and Wang, Yuqing and Lu, Chi and Lai, Christopher and He, Yanjun and Shao, Xun and Xie, Zhuoqing and others},
  journal={arXiv preprint arXiv:2410.08474},
  year={2024}
}

@article{liu2025f,
  title={${F} ^{3}$Set: Towards Analyzing Fast, Frequent, and Fine-grained Events from Videos},
  author={Liu, Zhaoyu and Jiang, Kan and Ma, Murong and Hou, Zhe and Lin, Yun and Dong, Jin Song},
  journal={arXiv preprint arXiv:2504.08222},
  year={2025}
}

@article{liu2023visual,
  title={Visual instruction tuning},
  author={Liu, Haotian and Li, Chunyuan and Wu, Qingyang and Lee, Yong Jae},
  journal={Advances in neural information processing systems},
  volume={36},
  pages={34892--34916},
  year={2023}
}

@article{zou2024seconds,
  title={From seconds to hours: Reviewing multimodal large language models on comprehensive long video understanding},
  author={Zou, Heqing and Luo, Tianze and Xie, Guiyang and Lv, Fengmao and Wang, Guangcong and Chen, Junyang and Wang, Zhuochen and Zhang, Hansheng and Zhang, Huaijian and others},
  journal={arXiv preprint arXiv:2409.18938},
  year={2024}
}

@article{maaz2023video,
  title={Video-chatgpt: Towards detailed video understanding via large vision and language models},
  author={Maaz, Muhammad and Rasheed, Hanoona and Khan, Salman and Khan, Fahad Shahbaz},
  journal={arXiv preprint arXiv:2306.05424},
  year={2023}
}

@article{lin2023video,
  title={Video-llava: Learning united visual representation by alignment before projection},
  author={Lin, Bin and Ye, Yang and Zhu, Bin and Cui, Jiaxi and Ning, Munan and Jin, Peng and Yuan, Li},
  journal={arXiv preprint arXiv:2311.10122},
  year={2023}
}

@article{bai2025qwen2,
  title={Qwen2. 5-vl technical report},
  author={Bai, Shuai and Chen, Keqin and Liu, Xuejing and Wang, Jialin and Ge, Wenbin and Song, Sibo and Dang, Kai and Wang, Peng and Wang, Shijie and Tang, Jun and others},
  journal={arXiv preprint arXiv:2502.13923},
  year={2025}
}

@article{ye2024mplug,
  title={mplug-owl3: Towards long image-sequence understanding in multi-modal large language models},
  author={Ye, Jiabo and Xu, Haiyang and Liu, Haowei and Hu, Anwen and Yan, Ming and Qian, Qi and Zhang, Ji and Huang, Fei and Zhou, Jingren},
  journal={arXiv preprint arXiv:2408.04840},
  year={2024}
}

@article{wang2025internvideo2,
  title={Internvideo2. 5: Empowering video mllms with long and rich context modeling},
  author={Wang, Yi and Li, Xinhao and Yan, Ziang and He, Yinan and Yu, Jiashuo and Zeng, Xiangyu and Wang, Chenting and Ma, Changlian and Huang, Haian and Gao, Jianfei and others},
  journal={arXiv preprint arXiv:2501.12386},
  year={2025}
}

@article{xiaomi2025mimo,
  title={MiMo: Unlocking the Reasoning Potential of Language Model--From Pretraining to Posttraining},
  author={Xiaomi, LLM and Xia, Bingquan and Shen, Bowen and Zhu, Dawei and Zhang, Di and Wang, Gang and Zhang, Hailin and Liu, Huaqiu and Xiao, Jiebao and Dong, Jinhao and others},
  journal={arXiv preprint arXiv:2505.07608},
  year={2025}
}

@article{feng2025video,
  title={Video-r1: Reinforcing video reasoning in mllms},
  author={Feng, Kaituo and Gong, Kaixiong and Li, Bohao and Guo, Zonghao and Wang, Yibing and Peng, Tianshuo and Wu, Junfei and Zhang, Xiaoying and Wang, Benyou and Yue, Xiangyu},
  journal={arXiv preprint arXiv:2503.21776},
  year={2025}
}

@article{li2025videochat,
  title={Videochat-r1: Enhancing spatio-temporal perception via reinforcement fine-tuning},
  author={Li, Xinhao and Yan, Ziang and Meng, Desen and Dong, Lu and Zeng, Xiangyu and He, Yinan and Wang, Yali and Qiao, Yu and Wang, Yi and Wang, Limin},
  journal={arXiv preprint arXiv:2504.06958},
  year={2025}
}

@article{wang2025videorft,
  title={VideoRFT: Incentivizing Video Reasoning Capability in MLLMs via Reinforced Fine-Tuning},
  author={Wang, Qi and Yu, Yanrui and Yuan, Ye and Mao, Rui and Zhou, Tianfei},
  journal={arXiv preprint arXiv:2505.12434},
  year={2025}
}

@article{chen2025versavid,
  title={VersaVid-R1: A Versatile Video Understanding and Reasoning Model from Question Answering to Captioning Tasks},
  author={Chen, Xinlong and Zhang, Yuanxing and Guan, Yushuo and Zeng, Bohan and Shi, Yang and Yang, Sihan and Wan, Pengfei and Liu, Qiang and Wang, Liang and Tan, Tieniu},
  journal={arXiv preprint arXiv:2506.09079},
  year={2025}
}

@article{guo2025deepseek,
  title={Deepseek-r1: Incentivizing reasoning capability in llms via reinforcement learning},
  author={Guo, Daya and Yang, Dejian and Zhang, Haowei and Song, Junxiao and Zhang, Ruoyu and Xu, Runxin and Zhu, Qihao and Ma, Shirong and Wang, Peiyi and Bi, Xiao and others},
  journal={arXiv preprint arXiv:2501.12948},
  year={2025}
}

@article{kojima2022large,
  title={Large language models are zero-shot reasoners},
  author={Kojima, Takeshi and Gu, Shixiang Shane and Reid, Machel and Matsuo, Yutaka and Iwasawa, Yusuke},
  journal={Advances in neural information processing systems},
  volume={35},
  pages={22199--22213},
  year={2022}
}
}
\end{document}